# Multilingual Text Analysis for Text-to-Speech Synthesis

Richard Sproat[1]

**Abstract.** We present a model of text analysis for text-to-speech (TTS) synthesis based on (weighted) finite-state transducers, which serves as the text-analysis module of the multilingual Bell Labs TTS system. The transducers are constructed using a lexical toolkit that allows declarative descriptions of lexicons, morphological rules, numeral-expansion rules, and phonological rules, inter alia. To date, the model has been applied to eight languages: Spanish, Italian, Romanian, French, German, Russian, Mandarin and Japanese.

## 1 Introduction

The first task faced by any text-to-speech (TTS) system is the conversion of input text into an internal linguistic representation. This is in general a complex task since the written form of any language is at best an imperfect representation of the corresponding spoken forms. Among the problems that one faces in handling ordinary text are the following:

1. While a large number of languages delimit words using whitespace or some other device, some languages, such as Chinese and Japanese do not. One is therefore required to 'reconstruct' word boundaries in TTS systems for these languages.
2. Digit sequences need to be expanded into words, and more generally into well-formed number names: so *243* in English would generally be expanded as *two hundred and forty three*.
3. Abbreviations need to be expanded into full words. In general this can involve some amount of contextual disambiguation: so *kg.* can be either *kilogram* or *kilograms*, depending upon the context.
4. Ordinary words and names need to pronounced. In many languages, this requires morphological analysis: even in languages with fairly 'regular' spelling, morphological structure is often crucial in determining the pronunciation of a word.
5. Prosodic phrasing is only sporadically indicated (by punctuation marks) in the input, and phrasal accentuation is almost never indicated. At a minimum, some amount of lexical analyis (in order to determine, e.g. grammatical part of speech) is necessary in order to predict which words to make prominent, and where to place prosodic boundaries.

In many TTS systems the first three tasks — word segmentation, and digit and abbreviation expansion — would be classed under the rubric of *text normalization* and would generally be handled prior to, and often in a quite different fashion from the last two problems, which fall more squarely within the domain of linguistic analysis[2]

One problem with this approach is that in many cases the selection of the correct linguistic form for a 'normalized' item cannot be chosen before one has done a certain amount of linguistic analysis. Consider an example that is problematic for the Bell Laboratories American English TTS system, a system that treats text normalization prior to, and separately from, the rest of linguistic analysis. If one encounters the string *$5* in an English text, the normal expansion would be *five dollars*. But this expansion is not always correct: when functioning as a prenominal modifier, as in the phrase *$5 bill*, the correct expansion is *five dollar*, since in general plural noun forms cannot function as modifiers in English. The analysis of complex noun phrases in the American English system (cf. [14]) comes later than the preprocessing phase, and since a hard decision has been made in the earlier phase, the system produces an incorrect result.

An even more compelling example can be found in Russian. While in English the percentage symbol '%', when denoting a percentage, is always read as *percent*, in Russian selecting the correct form depends on complex contextual factors. The first decision that needs to be made is whether or not the number-percent string is modifying a following noun. Russian in general disallows noun-noun modification: in constructions equivalent to noun-noun compounds in English, the first noun must be converted into an adjective: thus *rog* 'rye', but *rzhanoj xleb* (rye+adj bread) 'rye bread'. This general constraint applies equally to *procent* 'percent', so that the correct rendition of *20% skidka* 'twenty percent discount' is *dvadcati-procentnaja skidka* (twenty$_{[\text{gen}]}$-percent+adj$_{[\text{nom,sg,fem}]}$ discount$_{[\text{nom,sg,fem}]}$). Note that not only does *procent* have to be in the adjectival form, but as with any Russian adjective it must also agree in number, case and gender with the following noun. Observe also that the word for 'twenty' must occur in the genitive case. In general, numbers which modify adjectives in Russian must occur in the genitive case: consider, for example, *etazh* 'storey', and *dvux-etazhnyj* (two$_{[\text{gen}]}$-storey+adj$_{[\text{nom,sg,masc}]}$). If the percentage expression is not modifying a following noun, then the nominal form *procent* is used. However this form appears in different cases depending upon the number it occurs with. With numbers ending in *one* (including compound numbers like *twenty one*), *procent* occurs in the nominative singular. After so-called paucal numbers — *two*, *three*, *four* and their compounds — the genitive singular *procenta* is used. After all other numbers one finds the genitive plural *procentov*. So we have *odin procent* (one percent$_{[\text{nom,sg}]}$), *dva procenta* (two percent$_{[\text{gen,sg}]}$), and *pyat' procentov* (five percent$_{[\text{gen,pl}]}$). All of this, however, presumes that the percentage expression as a whole is in a non-oblique case. If the expression is in an oblique case, then both the number and *procent* show up in that case, with *procent* being in the singular if the number ends in *one*, and the plural otherwise: *s odnym procentom* (with one$_{[\text{instr,sg,masc}]}$ percent$_{[\text{instr,sg}]}$) 'with one percent'; *s pjat'ju procentami* (with five$_{[\text{instr,pl}]}$ percent$_{[\text{instr,pl}]}$) 'with five percent'. As with the adjectival forms, there is nothing peculiar about the behavior of the noun *procent*: all nouns exhibit similar behavior in combination with numbers (cf. [2]). The complexity, of course, arises because the written form '*%*' gives no indication of what linguistic form it corresponds to. Furthermore, there is no way to correctly expand this form without doing a substantial amount of analysis of the context, including some analysis of the morphological properties of the surrounding words, as well as an analysis of the relationship of the percentage expression to those words.

---

[1] Speech Synthesis Research Department, Bell Laboratories, Lucent Technologies, 700 Mountain Avenue, Room 2d-451, Murray Hill, NJ, USA, 07974–0636, rws@bell-labs.com

[2] But see [17], which treats numeral expansion as an instance of morphological analysis, and also the work of van Leeuwen [19], which uses cascaded rewrite rules within his Too$_\text{L}$iP system towards the same ends.

The obvious solution to this problem is to delay the decision on how exactly to transduce symbols like '$' in English or '%' in Russian until one has enough information to make the decision in an informed manner. This suggests a model where, say, an expression like '20%' in Russian is transduced into all possible renditions, and the correct form selected from the lattice of possibilities by filtering out the illegal forms. An obvious computational mechanism for accomplishing this is the *finite-state transducer* (FST). Indeed, since it is well-known that FSTs can also be used to model (most) morphology and phonology [9, 8, 13], as well as to segment words in Chinese text [16], and (as we shall argue below) for performing other text-analysis operations such as numeral expansion, this suggests a model of text-analysis that is entirely based on regular relations. We present such a model below. More specifically we present a model of text analysis for TTS based on *weighted* FSTs (WFSTs) [12, 11], which serves as the text-analysis module of the multilingual Bell Labs TTS system. To date, the model has been applied to eight languages: Spanish, Italian, Romanian, French, German, Russian, Mandarin and Japanese. One property of this model which distinguishes it from most text-analyzers used in TTS systems is that there is no sense in which such tasks as numeral expansion or word-segmentation are logically prior to other aspects of linguistic analysis, and there is therefore no distinguished 'text-normalization' phase.

## 2 Overall Architecture

Let us start with the example of the lexical analysis and pronunciation of ordinary words, taking again an example from Russian. Russian orthography is often described as 'morphological' [1], meaning that the orthography represents not a surface phonemic level of representation, but a more abstract level. This is description is correct, but from the point of view of predicting word pronunciation, it is noteworthy that Russian, with a well-defined set of lexical exceptions, is almost completely phonemic in that one can predict the pronunciation of most words in Russian based on the spelling of those words — provided one knows the placement of word stress, since several Russian vowels undergo reduction to varying degrees depending upon their position relative to stressed syllables. The catch is that word stress usually depends upon knowing lexical properties of the word, including morphological class information. To take a concrete example, consider the word *kostra* (Cyrillic костра) (bonfire+genitive.singular). This word belongs to a class of masculine nouns where the word stress is placed on the inflectional ending, where there is one. Thus the stress pattern is *kostr'a*, and the pronunciation is /kəstr'ʌ/, with the first /o/ reduced to /ə/. Let us assume that the morphological representation for this word is something like $kostr\{noun\}\{masc\}\{inan\}+'a\{sg\}\{gen\}$, where for convenience we represent phonological and morphosyntactic information as part of the same string.[3] Assuming a finite-state model of lexical structure [9, 8], we can easily imagine a set of transducers $M$ that map from that level into a level that gives the minimal morphologically-motivated annotation (MMA) necessary to pronounce the word. In this case, something like *kostr'a* would be appropriate. Call this *lexical-to-MMA* transducer $L_{word}$; such a transducer can be constructed by composing the lexical acceptor $D$ with $M$ so that $L_{word} = D \circ M$. A transducer that maps from the MMA to the standard spelling *kostra* (костра) would, among other things, simply delete the stress marks: call this transducer $S$. The composition $L_{word} \circ S$, then computes the mapping from the lexical level to the surface orthographic level, and its inverse $(L_{word} \circ S)^{-1} = S^{-1} \circ L_{word}^{-1}$ computes the mapping from the surface to all possible lexical representations for the text word. A set of pronunciation rules compiled into a transducer (cf. [5] and below) $P$, maps from the MMA to the (surface) phonological representation; note that by starting with the MMA, rather than with the more abstract lexical representation, the pronunciation rules do not need to duplicate information that is contained in $L_{word}$ anyway. Mapping from a single orthographic word to its pronunciation thus involves composing the acceptor representing the word with the transducer $S^{-1} \circ L_{word}^{-1} \circ L_{word} \circ P$ (or more fully as $S^{-1} \circ M^{-1} \circ D \circ M \circ P$).

For textual elements such as numbers, abbreviations, and special symbols such as '%', the model just presented seems less persuasive, because there is no aspect of a string, such as '25%' that indicates its pronunciation: such strings are purely *logographic* — some might even argue *ideographic* — representing nothing about the phonology of the words involved.[4] For these cases we presume a direct mapping between all possible forms of *procent*, and the symbol '%': call this transducer $L_{perc}$, a subset of $L_{special\ symbol}$. Then $L_{perc}^{-1}$ maps from the symbol '%' to the various forms of *procent*. In the same way, the transducer $L_{numbers}^{-1} \frown L_{perc}^{-1}$ maps from numbers followed by the sign for percent, into various possible (and some impossible) lexical renditions of that string — the various forms to be disambiguated using contextual information, as we shall show later on. Abbreviations are handled in a similar manner: note that abbreviations such as *kg* (кг) in Russian show the same complexity of behavior as *procent*.

So far we have been discussing the mapping of single text words into their lexical renditions. The construction of an analyzer to handle a whole text is based on the observation that a text is simply constructed out of zero or more instances of a text word coming from one of the models described above — i.e., either an ordinary word, an abbreviation, a number, a special symbol, or possibly some combination of numbers with a special symbol; with each of these tokens separated by some combination of whitespace or punctuation. The structure of this model of lexical analysis is summarized in Figure 1.

We presume two models for space and punctuation. The model $L_{\text{SPACE}}^{-1}$, maps between interword SPACE and its potential lexical realizations, usually a word boundary, but in some cases a higher-level prosodic phrase boundary. Interword SPACE is parameterized so that in European languages, for example, it corresponds to actual whitespace, whereas in Chinese or Japanese, it corresponds to $\epsilon$. Similarly, the model $L_{punc}^{-1}$ maps between punctuation marks (possibly with flanking whitespace) and the lexical realization of those marks: in many, though not all, cases the punctuation mark may correspond to a prosodic phrase boundary.

The output of the lexical analysis WFST diagramed in Figure 1 is a lattice of all possible lexical analyses of all words in the input sentence. Obviously in general we want to remove contextually inappopriate analyses, and to pick the 'best' analysis in cases where one cannot make a categorical decision. This is accomplished by a set of one or more *language model* transducers — henceforth Λ — which are derived from rules and other linguistic descriptions that apply to contexts wider than the lexical word. Phrasal accentuation and prosodic phrasing are also handled by the language model transducers.[5] The output of composing the lexical analysis WFST with Λ is a lattice of contextually disambiguated lexical analyses. The best-path of this lattice is then selected, using a Viterbi *best path* algorithm. Costs on the lattice may be costs hand-selected to disfavor certain lexical analyses — see the Russian percentage example detailed in a subsequent section; or they may be genuine data-derived cost estimates, as in the case of the Chinese lexical analysis WFST, where the costs correspond to the negative log (unigram) probability of a particular lexical entry [16]. Given the best lexical analysis, one can then proceed to apply the phonological transducer (or set of transducers) $P$ to the lexical analysis, or more properly to the lexical analysis composed with the lexical-to-MMA map $M$, as we saw above. Although the lexical-to-MMA map $M$ was introduced as mapping from the lexical analyses of ordinary words to their MMA, if the map is constructed with sufficient care it can serve as the transducer for lexical analyses coming from any of the text-word models.

---

[3] Conversion between a 'flattened' representation of this kind and a hierarchical representation more in line with standard linguistic models of morphology and phonology is straightforward and we will not dwell on this issue here.

[4] This is certainly not completely true in all such cases, as 'mixed' representations such as *1st* and *2nd* suggest. But such cases are most easily treated as also being logographic, at least in the present architecture.

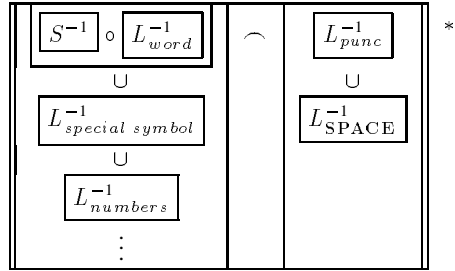

**Figure 1.** Overall structure of the lexical analysis portion. Note that $SPACE$ corresponds to whitespace in German or Russian, but $\epsilon$ in Chinese or Japanese.

## 3 The Tools

The construction of the WFSTs depends upon a lexical toolkit — *lextools* — that allows one to describe linguistic generalizations in linguistically sensible human-readable form. The toolkit has more or less the same descriptive power as the Xerox tools [7, 6], though the current version of lextools lacks some of the debugging capabilities of the Xerox system, and the Xerox tools do not allow costs in the descriptions whereas lextools does.[6]

Some of the tools do not require much comment for readers familiar with previous work on finite-state phonology and morphology. In addition to some basic tools to deal with machine labels, there is a tool (*compwl*) that compiles lists of strings or more general regular expressions into finite-state lexicons; a tool (*paradigm*) to construct inflected morphological forms out of inflectional paradigm descriptions and lexicons that mark the paradigm affiliation of stems; a tool for constructing finite-state word grammars (*arclist* — the name being inspired by [18]); and a rewrite rule compiler [5], based on the algorithm described in [10].

The tool *numbuilder* constructs the transducer $L^{-1}_{numbers}$, which converts strings of digits up to a user-defined length into number-names appropriate for that string. The construction factors the problem of numeral expansion into two subproblems. The first of these involves expanding the numeral sequence into a representation in terms of sums of products of powers of the base — usually ten; call this the *Factorization* transducer. The second maps from this representation into number names using a lexicon that describes the mapping between basic number words and their semantic value in terms of sums of products of powers of the base; call this the *Number Lexicon* transducer. A sample Number Lexicon fragment for German is shown in Figure 2. The first stage of the expansion is language- or at least language-area dependent since languages differ on which powers of the base have separate words (see [4], inter alia): so Chinese and Japanese have a separate word for $10^4$, whereas most European languages lack such a word. The full numeral expansion transducer is constructed by composing the Factorization transducer with the transitive closure of the Number Lexicon transducer. In some languages, additional manipulations are necessary, and these involve the insertion of a special 'filter' transducer between the Factorization and Number Lexicon transducers. In German, for example, the words for decades come after the words for units: so $34 = 3 \cdot 10^1 + 4$ becomes *vierunddreißig* (four+and+thirty), thus suggesting the order $4 + 3 \cdot 10^1$. This reversal can be accomplished by a filter transducer, which for lack of a better name we will call *Decade Flop*. The construction of the numeral expander for German is shown schematically in Figure 3. (Note that the insertion of *und* in examples like *vierunddreißig* 'thirty four' is not actually accomplished by numbuilder: such 'clean up' operations can be handled using rewrite rules.)

One tool that has so far not been applied in the system is the decision tree compiler, described in [15]. We are hoping to apply this in the future for, among other things, phrasing models of the kind discussed in [20].

## 4 Russian Percentages

Let us return to the example of Russian percentage terms. Assume that we start with a fragment of text such as с 5% скидкой *s 5% skidkoj* (with 5% discount) 'with a five-percent discount'. This is first composed with the lexical analysis WFST to produce a set of possible

---

[5] To date our multilingual systems have rather rudimentary lexical-class based accentuation rules, and punctuation-based phrasing. Thus these components of the systems are not as sophisticated as the equivalent components of our English system [20, 3, 14]. This is largely because the relevant research has not been done for most of the languages in question, rather than for technical problems in fitting the results of that research into the model.

[6] The work is also similar to the Too$_L$iP toolkit for linguistic rules for TTS discussed in [19]. However, the latter does not compile the rules into (W)FSTs. Instead, the right and left contexts are compiled into an FSM-like format, which is then used to match these contexts for the rules at runtime. Note also that, unlike the current system which compiles linguistic descriptions into WFSTs that allow multiple outputs, Too$_L$iP functions in a completely deterministic fashion, in that for any given input there can only be one output.

```
/{1}                    :   ('eins{num}({masc}|{neut}){sg}{##})/
/{1}                    :   ('eine{num}{femi}{sg}<1.0>{##})/
/{2}                    :   (zw'ei{num}{##})/
/{3}                    :   (dr'ei{num}{##})/
                            ⋮
/({0}{+++}{1}{10∧1})    :   (z'ehn{num}{##})/
/({1}{+++}{1}{10∧1})    :   ('elf{num}{##})/
/({2}{+++}{1}{10∧1})    :   (zw'ölf{num}{##})/
/({3}{+++}{1}{10∧1})    :   (dr'ei{++}zehn{num}{##})/
                            ⋮
/({2}{10∧1})            :   (zw'an{++}zig{num}{##})/
/({3}{10∧1})            :   (dr'ei{++}ßig{num}{##})/
                            ⋮
/({10∧2})               :   (h'undert{num}{##})/
/({10∧3})               :   (t'ausend{num}{neut}{##})/
```

**Figure 2.** German number lexicon. The cost of 1.0 on the feminine form *eine* effectively disfavors this form, so that it will only be selected in appropriate predefined contexts.

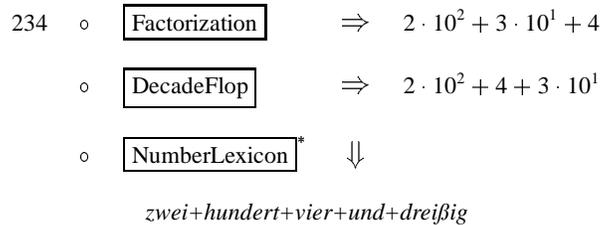

**Figure 3.** Expansion of *234* in German using *numbuilder*.

lexical forms; see Figure 4. By default the lexical analyzer marks the adjectival readings of '%' with '⋆', meaning that they will be filtered out by the language-model WFSTs, if contextual information does not save them. Costs on analyses (here represented as subscripted floating-point numbers) mark constructions — usually oblique case forms — that are not in principle ill-formed but are disfavored except in certain well-defined contexts. The correct analysis (boxed in Figure 4), for example, has a cost of 2.0 which is an arbitrary cost assigned to the oblique instrumental adjectival case form: the preferred form of the adjectival rendition '%' is masculine, nominative, singular if no constraints apply to rule it out.

Next the language model WFSTs $\Lambda$ are composed with the lexical analysis lattice. The WFSTs $\Lambda$ include transducers compiled from rewrite rules that ensure that the adjectival rendition of '%' is selected whenever there is a noun following the percent expression, and rules that ensure the correct case, number and gender of the adjectival form given the form of the following noun. In addition, a filter expressable as $\neg(\Sigma^* \boxed{\star} \Sigma^*)$ removes any analyses containing the tag '⋆'. See Figure 5. The best-cost analysis among the remaining analyses is then selected. Finally, the lexical analysis is composed with $M \circ P$ to produce the phonemic transcription; see Figure 6.

## 5 Size and Speed Issues

Table 1 gives the sizes of the lexical analysis WFSTs for the languages German, Spanish, Russian and Mandarin. To a large extent, these sizes accord with our intuitions of the difficulties of lexical processing in the various languages. So Russian is very large, correlating with the complexity of the morphology in that language. German is somewhat smaller. Mandarin has a small number of states, correlating with the fact that Mandarin words tend to be simple in terms of morphemic structure; but there are a relatively large number of arcs, due to the large character set involved. Sizes for the Spanish transducer are misleading since the current Spanish system includes only minimal morphological analysis: note, though, that morphological analysis is mostly unnecessary in Spanish for correct word pronunciation.

While the transducers can be large, the performance (on an SGI Indy or Indigo) is acceptably fast for a TTS application. Slower performance is certainly observed, however, when the system is required to explore certain areas of the network, as for example in the case of expanding and disambiguating Russian number expressions.

To date, no formal evaluations have been performed on the correctness of word-pronunciation in the various languages we are working on, largely because there is still work to be done before the systems can be called complete. An evaluation of the correctness of word segmentation in the Mandarin Chinese system is reported in [16].

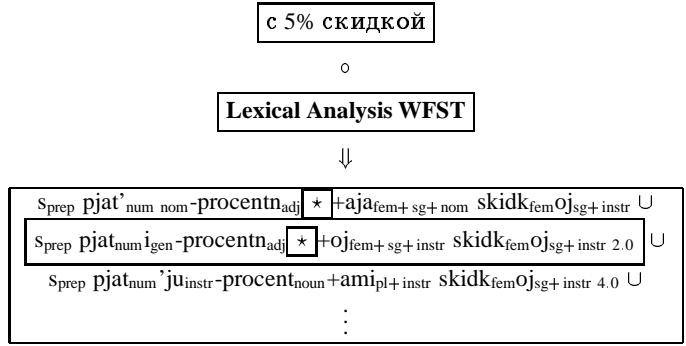

**Figure 4.** Composition of с 5% скидкой *s 5% skidkoj* 'with a 5% discount' with the lexical analysis WFST to produce a range of possible lexical renditions for the phrase. By default the adjectival readings of '%' are marked with '⋆', which means that they will be filtered out by the language-model WFSTs; see Figure 5. The boxed analysis is the correct one. Costs on analyses mark constructions — usually oblique case forms — that are not in principle ill-formed but are disfavored except in certain well-defined contexts.

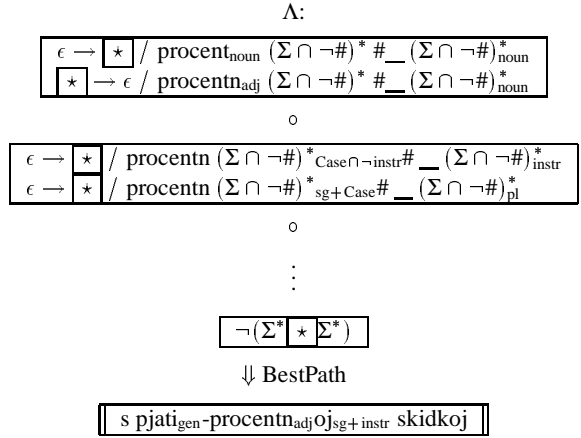

**Figure 5.** A subset of the language model WFSTs related to the rendition of percentages. The first block of rules ensures that adjectival forms are used before nouns, by switching the tag '⋆' on the adjectival and nominal forms. The second block of rules deals with adjectival agreement with the adjectival forms. The final block is a filter ruling out the forms tagged with '⋆'. The (correct) output of this sequence of transductions is shown at the bottom.

|  | States | Arcs |
|---|---|---|
| German | 77295 | 207859 |
| Russian | 139592 | 495847 |
| Mandarin | 48015 | 278905 |
| Spanish | 8602 | 17236 |

**Table 1.** Sizes of lexical analysis WFSTs for selected languages.

## 6 Summary and Future Work

The system for text analysis presented in this paper is a complete working system that has been used in the development of working text-analysis systems for several languages. In addition to German, Spanish, Russian and Mandarin, a system for Romanian has been built, and work on French, Japanese and Italian is underway. From the point of view of previous research on linguistic applications finite-state transducers, some aspects of this work are familiar, some less so. Familiar, of course, are applications to morphology, phonology, and syntax, though most previous work in these areas has not made use of *weighted* automata. More novel are the applications to text 'preprocessing', in particular numeral expansion and word segmentation.

From the point of view of text-analysis models for text-to-speech the approach is quite novel since, as described in the introduction, most previous work treats certain operations, such as word segmentation or numeral expansion in a preprocessing phase that is logically prior to the linguistic analysis phase; we have argued here against this view.

Two areas of future work both depend upon an important property of the FSM toolkit on top of which the lextools toolkit is built. Underlying the notion of an FSM is the more general notion of a *generalized* state machine (GSM). An important property of GSMs is that it is not necessary to know beforehand which arcs leave a

79

R. Sproat

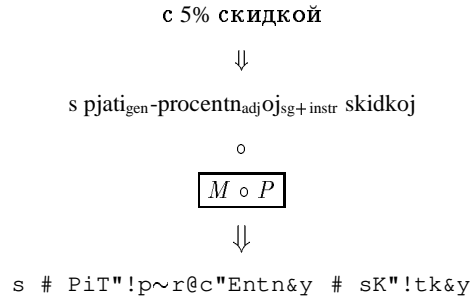

**Figure 6.** Mapping the selected lexical analysis to a phonetic rendition via composition with $P$.

given state; rather one can construct just the arcs one needs 'on the fly' as one is using the machine, for example in a composition with another machine. This has two important consequences. First of all, for a strictly *finite* state machine, it is not necessary to explicitly construct the machine beforehand, and this in turn implies that one can avoid precompiling very large FSMs, so long as one can provide an algorithm for constructing the machine on the fly. One example is in discourse analysis, where one wants to remember which words or lemmata one has already seen; as previous work on accenting suggests [3], this kind of information is useful for TTS, and is in fact used in the American English version of the Bell Labs synthesizer. In theory, assuming the set of words or morphological stems is closed, one could construct an FSM that would 'remember' when it had seen a word; needless to say, such a machine would be astronomical in size, and so the precompilation of this machine is out of the question; one could however envision dynamically building states that 'remember' that a particular word has been seen. Secondly, one can in principle construct GSMs which have greater than finite-state power, again providing that one can specify an algorithm for constructing on the fly the arcs leaving a given state. One obvious example is a 'copy machine' which will recognize which strings from a lattice have the property that they are of the form $ww$, for some string $w$; this problem comes up in the analysis of morphological reduplication. Precompiling such a machine as an FSM for copies of unbounded length is of course impossible; however, it is possible to construct a GSM which can be composed with an arbitrary (acyclic) lattice and will find exactly those strings with the desired property. Future work on the text analysis model presented here will focus in part on the application of generalized state machines to various linguistic problems.

## 7 Acknowledgments

I wish to acknowledge Michael Riley, Fernando Pereira and Mehryar Mohri of AT&T Research, without whose underlying FSM toolkit the lexical toolkit and text-analysis work reported here would not have been possible.